# Vision Language Models for Primary Structure Threat Identification in Wildland Urban Interface Fires

Zhangding Liu
School of Computational Science and Engineering
Georgia Institute of Technology
Atlanta, GA, USA
zliu952@gatech.edu

Neda Mohammadi
School of Civil and Environmental Engineering
Georgia Institute of Technology
Atlanta, GA, USA
nedam@gatech.edu
The University of Sydney
Sydney, NSW, Australia
neda.mohammadi@sydney.edu.au

John E. Taylor
School of Civil and Environmental Engineering
Georgia Institute of Technology
Atlanta, GA, USA
jet@gatech.edu

*Abstract*—**Wildland urban interface (WUI) fires increasingly threaten communities and infrastructure, motivating the need for rapid visual assessment. Recent research has focused primarily on detecting fire and assessing post-event damage. However, structure-level fire threat identification during active WUI fire events remains underexplored. This study evaluates Qwen3-VL-8B-Instruct for this active-fire decision using two separately scored tracks: 1,824 synthetic images from 491 designed WUI scenarios and 45 frames from a controlled-burn experiment conducted in collaboration with Columbus Fire & EMS Department. Under zero-shot prompting, the model returned valid target fields for 1,808 synthetic images and 39 controlled-burn frames. Accuracy among valid outputs was 94.0% for the synthetic track and 69.2% for the controlled-burn track, compared with a 16.0% random-selection baseline for the latter. On the synthetic track, the model generally identified the intended target across varied designed WUI scenarios when the relevant fire-spread cues were visually distinct. Controlled-burn performance was well above random selection, but errors showed difficulty interpreting spatial relations in complex scenes. The dual-track protocol provides scalable stress testing in designed scenarios and a controlled-burn reference for assessing model behavior under physical fire conditions. Future work can use the synthetic data for VLM post-training and then evaluate the adapted models on larger and more diverse real WUI fire datasets.**

## I. INTRODUCTION

In April 2026, the Highway 82 Fire in Brantley County, Georgia, destroyed 87 homes and forced thousands of residents to evacuate after a foil balloon contacted a power line and ignited dry vegetation [1]. Such events are not limited to the fire prone western United States: when the 2018 Camp Fire reached Paradise, California, it spread into wildland urban interface (WUI) communities within hours, destroying thousands of structures as more than 30,000 residents evacuated [2]. The WUI, where developed land meets wildland vegetation, now covers roughly 5% of global land area and is home to more than one billion people [3]. Fast-moving incidents like these require commanders to make time-critical judgments about primary structure exposure, defensibility, and evacuation priorities [4]. These judgments precede any defensive action and must be made from the visible scene. WUI exposure is growing in both eastern and western regions, increasing the need for tools that support structure level threat assessment.

Camera-based wildfire systems have advanced fire detection and localization. Deployed systems such as ALERTCalifornia [5] and FUEGO [6] show that camera networks and remote sensing can identify the presence and location of fire. Vision language models (VLMs), which take an image and a text prompt as input and return a text answer, have begun to move beyond detection. Fine-tuned variants describe fire scenes in structured form [7], while studies of post-event damage show that single-view VLMs classify WUI building damage poorly and benefit from multi-view fusion [8]. Related disaster-assessment models include multi-label classifiers for hurricane damage [9] and a zero-shot VLM combined with a knowledge graph to estimate urban flood depth [10]. These efforts address disaster detection, description, or post-event assessment. However, structure-level threat assessment during WUI fires remains underexplored, particularly for identifying structures at immediate risk from an advancing fire. Primary threat identification requires comparing visible structures against the current fire, fuel, and barrier conditions. Such assessments could contribute to WUI fire digital twins for operator decision support [11], [12]. Generative models are increasingly used to populate these systems with data, scenarios, and 3D content [13].

Real controlled burn observations are valuable but limited, and collecting labeled WUI fire data at scale requires field access, fire events or burn experiments, and frame-level labeling. Controllable synthetic scenes offer a complementary route. Prior controlled-evaluation work in visual reasoning pairs rendered scenes with known labels to limit annotation noise [14]. Synthetic image generation has likewise been used to augment scarce training data in built-environment tasks [15]. Because a synthetic WUI scene starts from a designed layout, its fire location, structures, fuel paths, barriers, and intended highest-threat label are fixed in that layout before image generation. These scenes can expand scarce real-burn observations into scalable test settings, while comparison with controlled-burn frames remains necessary to assess real scene reliability.

This paper evaluates the zero-shot capability of Qwen3-VL-8B-Instruct for primary threat identification, pairing 1,808 synthetic images from 491 controllable scenes with 39 frames from a WUI controlled-burn experiment. The model performs well on synthetic scenes (94.0%) but less reliably on real-burn frames (69.2%), where accuracy shifts with the camera source. The gap between designed and real imagery motivates domain-specific validation before such models support smart-city wildfire response.

This paper makes three contributions:

- A controllable synthetic WUI scene pipeline that deliberately varies fire location, candidate structures, fuel paths, barriers, and viewpoint to produce diverse labeled evaluation scenes at scale.
- A dual-track evaluation that pairs scalable synthetic WUI fire scenes with scarce real controlled burn observations across operational camera viewpoints.
- A zero-shot diagnostic of Qwen3-VL-8B-Instruct [16] showing how much of its accuracy on designed scenes carries over to a real burn, and how much it shifts with capture conditions.

## II. METHODOLOGY

The evaluation centers on a structure-level decision in active WUI fire scenes: identifying the visible structure most immediately exposed to an advancing fire. The study uses two separately evaluated data sources. The first is a set of controllable synthetic WUI fire scenes generated from designed layouts. The second is a set of controlled burn frames collected from real fire experiments conducted in collaboration with the Columbus Fire & EMS Department in Columbus, GA, USA. Both sources are evaluated for the same primary threat identification task.

### A. *Synthetic WUI Scene Construction*

Each synthetic scene starts from a layout that specifies the intended fire location, candidate structures, structure roles, fuel paths, barriers, and intended primary-exposure label. This layout-first design gives each scene an explicit evaluation target before image generation. Fig. 1 summarizes this construction pipeline.

A structured image-generation prompt is derived from each layout. The prompt names the scene anchors, specifies the decision-relevant visual cues, fixes camera and composition constraints, and excludes confounding elements such as unintended roads, extra fences, or extra structures. Multi-structure scenes also require visible letter labels so that a model prediction can be matched to a specific target.

For each layout, the image generator returns several candidate images, and the pipeline admits them automatically for evaluation. Generative image models do not guarantee exact spatial fidelity, so a barrier may render wider or narrower than intended, fuel continuity may become visually ambiguous, or extra objects may appear. To check that automatic admission does not inflate accuracy, we separately validated a curated subset of 16 scenes and 78 images in which every image was human-reviewed.

### B. *VLM Prompting and Output Schema*

For each image, the VLM receives a fire threat reasoning prompt and returns a structured response. The evaluated field is primary_exposure, defined as the visible structure most at risk from the current fire, fuel, and barrier conditions. For controlled-burn frames, the model is instructed to act as a WUI structure triage analyst and assess every visible structure or human-made asset. Selected instructions include: "Choose the structure with the most direct, unobstructed fire pathway and highest threat." For barriers, it states: "Only mark a barrier as present if a visible non-combustible interruption lies between the fire and that structure." If the scene is unanswerable, the prompt requires abstention and a null primary_exposure.

The schema also records supporting information. The fuel_pathway field describes whether combustible material connects the fire to a structure. The barrier_present, barrier_type, and barrier_effectiveness fields describe physical features that may interrupt fire spread, such as fences, roads, cleared areas, or firelines. Per-structure fields describe the model's threat assessment for each candidate structure. These fields help diagnose whether the model ignored a barrier, over-weighted proximity, or treated an already-burned asset as still threatened.

### C. *Evaluation and Scoring Procedure*

Accuracy is computed by comparing the predicted primary exposure with the corresponding ground truth for each data source, and the two tracks are scored separately. For synthetic scenes, the prediction is compared with the layout-specified target label after automatic admission. For the controlled-burn frames, ground truth follows the physical experiment's on-site record of how the fire advanced; the authors then score each prediction as correct or incorrect for primary threat identification.

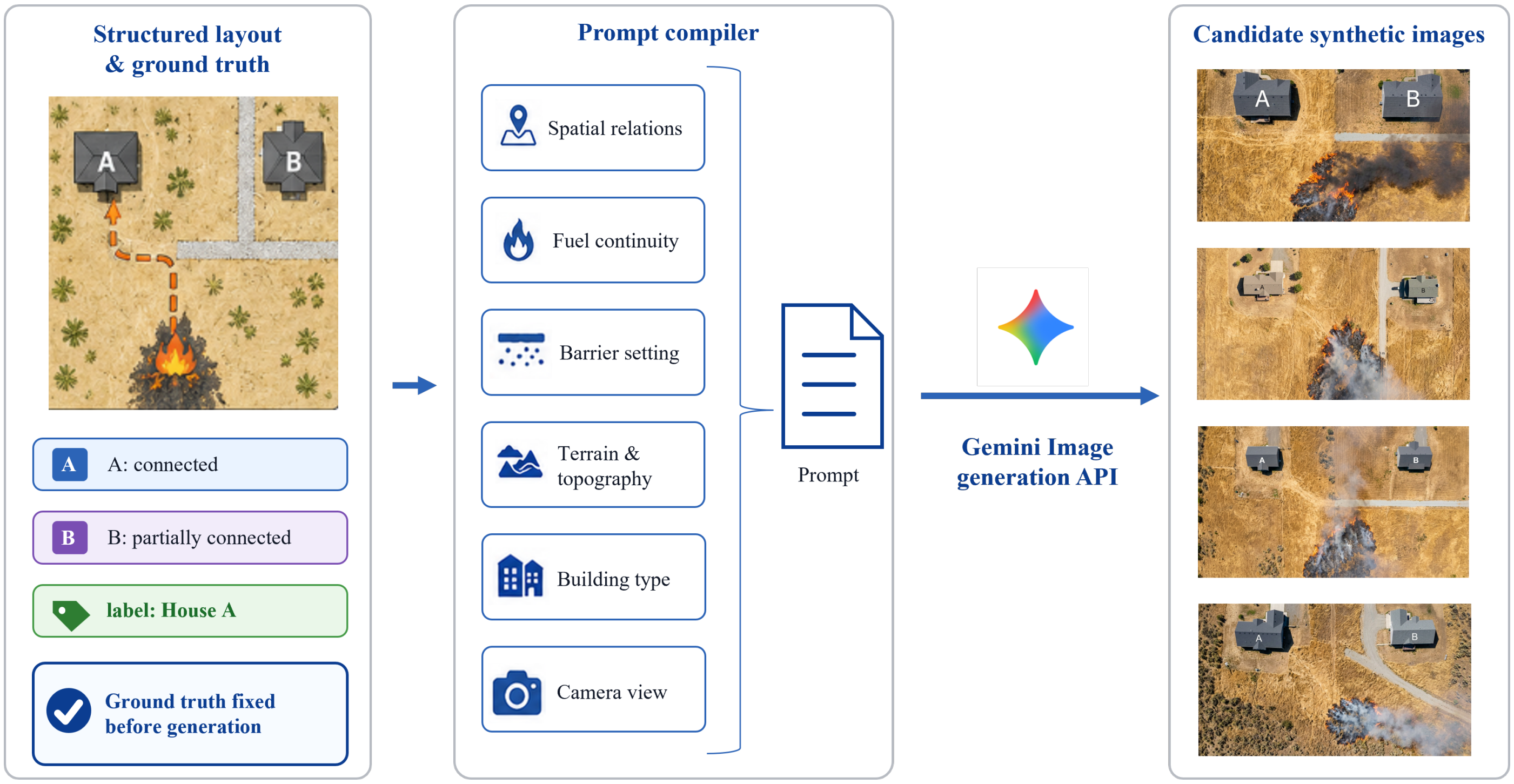

Fig. 1. Synthetic WUI scene construction pipeline. Each scene begins as a structured layout that fixes the intended target label before image generation. A prompt compiler converts the layout into an image generation prompt that specifies spatial relations, fuel continuity, barriers, terrain, building type, and camera view. The generative image model gemini-3.1-flash-image-preview [17] then produces candidate images.

## III. EXPERIMENTS

### A. *Synthetic Dataset*

The synthetic dataset contains 491 WUI fire scenes and 1,824 generated images, with multiple images per scene. Within this set, 16 scenes (78 images) form a per-image human-reviewed subset. The scenes were constructed to cover distinct primary threat identification conditions. They vary fire locations, candidate structures, fuel paths, and barriers. Across the set, the scenes span connected and interrupted fuel paths, fences and fuel breaks, multi-structure comparisons, special targets such as vehicles and manufactured home parks, and smoke-occluded views. Fig. 2 (top row) shows four example synthetic scenes.

### B. *Controlled-Burn Dataset*

The controlled-burn dataset consists of real video frames from a physical WUI fire experiment. We sampled 45 frames from 15 scenes across three video sources. Each frame is scored for the same primary threat identification task used in the synthetic track. The three sources provide different viewing conditions, and Fig. 2 (bottom row) shows an example frame from each source. Source A contains 14 frames from a ground-level video camera (JVC recorder, 1080p) positioned close to the burn, with wooden structures, burn piles, and clear fire-to-structure spatial geometry. Source B uses a ground-level smartphone (iPhone 14), with 14 frames where foreground and background structures are separated by a metal fence. Source C contains 11 frames from an elevated stationary camera with a wide overhead field of view, including trees, distant structures, and already-burned vehicles. Ground truth for the controlled-burn frames follows the physical experiment's on-site record of how the fire advanced, scored by the authors as correct or incorrect for primary threat identification. For the overhead source, the label refers to the most threatened unburned structure visible in the wide angle frame. The controlled burn set is small, and frames within a scene are correlated.

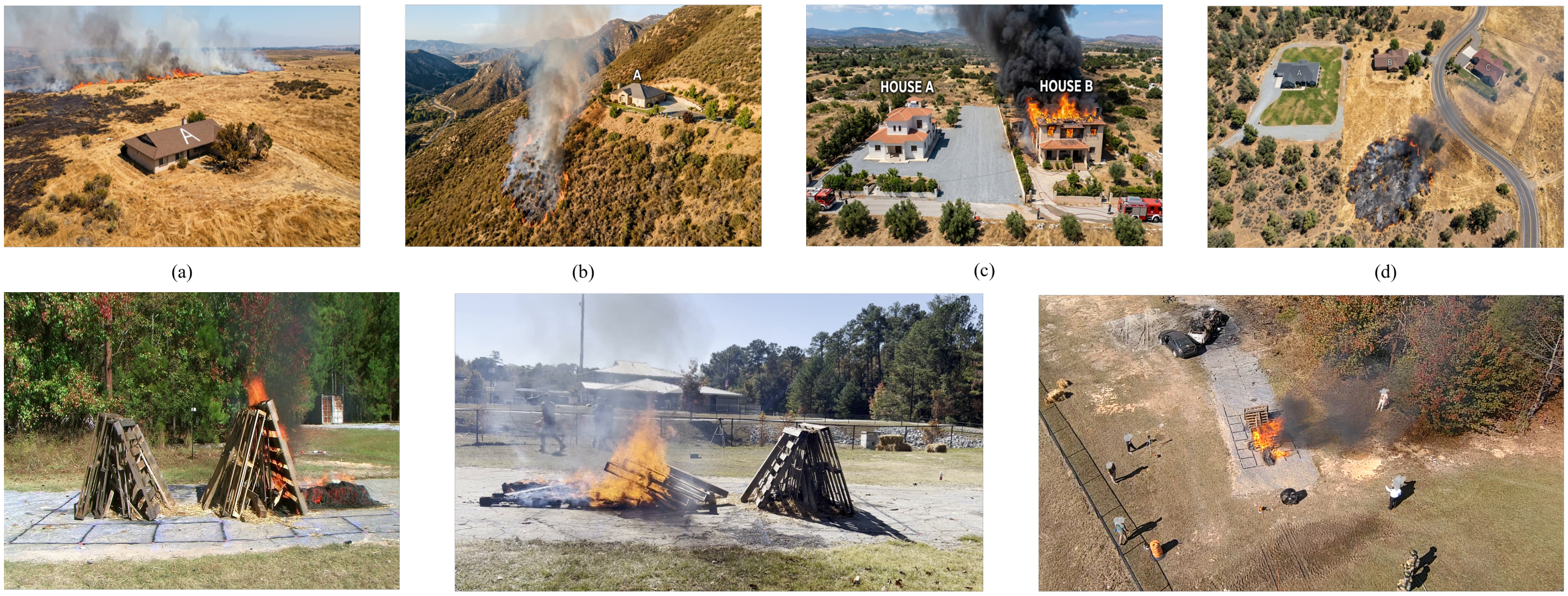

(a) (b) (c) (d)

(e) (f) (g)

Fig. 2. Example frames from both evaluation tracks, with correct primary_exposure given in parentheses. The top row shows synthetic scenes: (a) a continuous dry-grass pathway to a single house (House A), (b) uphill chaparral carrying fire up a slope (House A), (c) a structure-to-structure case in which a non-combustible gap protects one candidate while House A stays exposed (House A), and (d) a multi-target scene in which only one of three houses has an unobstructed pathway (House B). The bottom row shows one representative controlled-burn frame from each of the three capture sources: (e) a ground-level camera, (f) a ground-level smartphone captured from a different viewpoint, and (g) an elevated overhead stationary camera. In each controlled-burn frame, the primary_exposure is the wooden pallet directly adjacent to the currently burning pallet.

### C. *Setup*

Qwen3-VL-8B-Instruct [16] was used for fire threat reasoning evaluation, run on a single NVIDIA A100 GPU. The synthetic dataset uses the structured scene schema and is scored automatically against the layout-specified highest threat labels. The controlled-burn dataset is scored by manual binary judgment for each frame.

## IV. RESULTS

Table I reports primary threat identification accuracy for the synthetic and controlled-burn evaluations. Random selection is the expected accuracy of choosing uniformly among the candidate exposures enumerated in each frame. It is not computed for the synthetic sets, whose candidate structures vary by scene.

TABLE I. PRIMARY THREAT IDENTIFICATION ACCURACY FOR QWEN3-VL-8B-INSTRUCT UNDER ZERO-SHOT PROMPTING.

| **Evaluation set** | **n** | **Correct** | **Accuracy** | **Random selection** |
|---|---|---|---|---|
| Synthetic - curated subset | 78 | 73 | 93.6% | — |
| Synthetic - full set | 1808 | 1700 | 94.0% | — |
| Controlled burn - Source A (ground-level camera) | 14 | 12 | 85.7% | 16.7% |
| Controlled burn - Source B (ground-level smartphone) | 14 | 7 | 50.0% | 16.7% |
| Controlled burn - Source C (overhead stationary) | 11 | 8 | 72.7% | 14.3% |
| Controlled burn - overall | 39 | 27 | 69.2% | 16.0% |

### A. *Synthetic Evaluation*

On the synthetic set, 1,808 of the 1,824 generated images returned a valid primary-exposure field, and the model identified the most-threatened structure in 1,700 of them, corresponding to 94.0% accuracy. The model can perform primary threat identification well when scenes carry visually legible fire, fuel, barrier, and structure cues. Within the full auto-admitted set, the human-reviewed subset of 16 scenes and 78 images gives 93.6% accuracy. The close agreement between the subset and the full set indicates that automatic admission does not inflate accuracy and that the result holds as the set scales from 78 to 1,824 images.

The errors concentrate in structure-to-structure, woodpile, and distractor scenes, where the model selects a competing shed or returns None. The fuel_pathway field is the least stable, since the model frequently collapses both fully connected and fully blocked layouts into the conservative "partially connected" label when it meets visual gaps, roads, shadows, smoke, or oblique viewpoints. For barrier_present, it often reports a barrier from empty ground, road-like textures, surface changes, or the simple separation between structures.

### B. *Controlled-Burn Evaluation*

On the controlled-burn frames, 39 of the 45 sampled frames returned a valid primary-exposure field, and the model identifies the most-threatened structure in 27 of 39 frames, corresponding to 69.2% accuracy. For each frame, we enumerate the candidate exposures: two wood pallet stacks, a background structure or parked vehicles, two or three grass areas surrounding the fire, and the treeline. This yields six candidates for the two ground-level sources and seven for the overhead source. Under uniform random selection among these candidates, the baseline accuracy is 16.0%. In comparison, Qwen3-VL-8B-Instruct's accuracy of 69.2% is more than four times this baseline. Accuracy also varies across the three sources, highest for the close-range ground camera and lowest for the handheld phone. The sources differ in distance, device, field of view, occlusion, and background clutter.

Fig. 3 shows representative controlled-burn frames from the three capture sources, along with the model's primary-exposure response for each image. In Source A, the model selects the correct pallet when the threatened structure is visually salient, shifting from "wooden structure on the right" early in the burn to "pile of pallets on left" after the fire advances. It errs mainly when it names a background structure, as in Fig. 3(b), where the background building appears close to the flames in the 2D frame. The choice is visually understandable, but it reveals a failure to infer the true 3D separation between the fire and that structure. Source B introduces the opposite ambiguity from the phone viewpoint, where the metal fence drives failures such as "building behind fence," "building in background," and null predictions. In Source C, the overhead view captures the broader layout, but the model sometimes shifts to already burned vehicles, producing outputs such as "burned car near forest edge" instead of the intended pallet. The model's intermediate output shows that it treats the two adjacent pallets as a single combustible mass and then searches for a separate exposed target, which explains the choice.

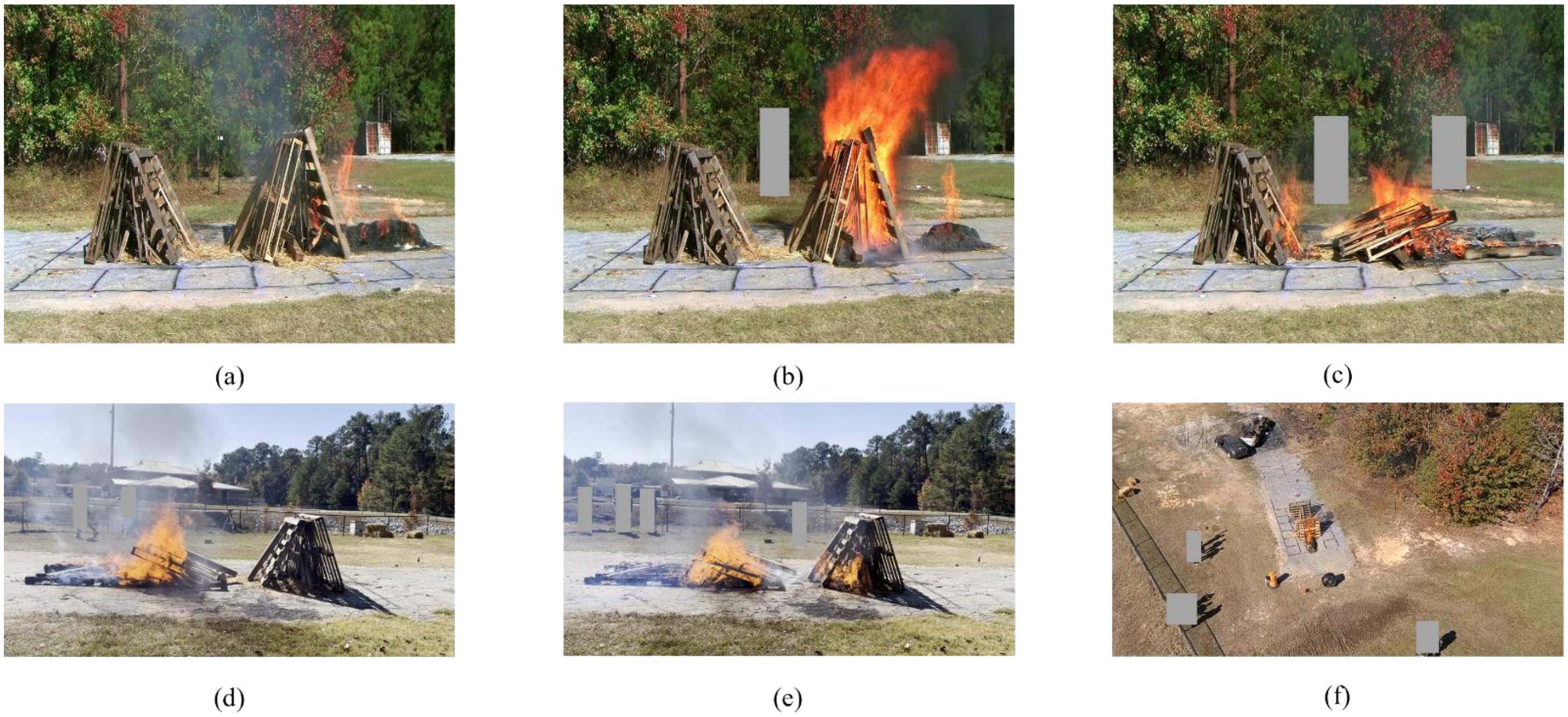

Fig. 3. Representative controlled-burn frames from the three capture sources, with the model's primary-exposure answer for each. Frames span an approximately 18-minute burn, pairing each stage of fire spread with a viewpoint. (a-c) Source A, as the fire moves from the ignited grass pile to the right pallet structure and then toward the left: the model answers "wooden structure on the right," then names the background building, then "pile of pallets on left." (d-e) Source B views the same burn from an opposing angle, where the metal fence produces "building behind fence." (f) Source C, overhead, where the model shifts to "burned car near forest edge." Boxes mask people for privacy.

## V. DISCUSSION

The synthetic and controlled-burn tracks evaluate the same primary-threat identification task under different data-acquisition conditions and dataset sizes. Among valid outputs, Qwen3-VL-8B-Instruct achieved 94.0% accuracy on synthetic images and 69.2% on controlled-burn frames, compared with a 16.0% random-selection baseline for the latter. The lower accuracy on controlled-burn frames shows that the task is more challenging in real-fire imagery. However, the difference between the two results also reflects differences in scene construction. Synthetic scenes are generated from layouts that specify the intended label and cues, whereas controlled-burn frames offer no comparable control over cue visibility or scene composition and are limited in sample size. The central implication is data-centric. Real controlled burn observations are valuable because they expose the model to actual flame, smoke, occlusion, and cluttered backgrounds, but they are scarce: the evaluated set contains 39 frames from one controlled-burn experiment and three camera sources. Controllable synthetic scenes help fill this gap by varying fire locations, structures, fuel paths, barriers, and viewpoints. This makes them useful for stress-testing primary threat identification before additional real burns are collected. The same logic underlies smart-city disaster digital twins [11], [12], where generated scenarios extend limited observations into repeatable test settings [13].

Despite the promising results, this study is still subject to several limitations. First, we evaluate only one open-weight 8B model under zero-shot prompting. Therefore, the reported results mainly reflect the performance of Qwen3-VL-8B-Instruct, and domain-specific adaptation may lead to different performance. Second, the controlled-burn result contains only 39 frames from a single burn, and frames within the same scene are correlated. The limited sample size reduces the precision of the per-source estimates, and the generalizability of the results to other building types, vegetation conditions, and weather conditions remains untested. Third, the synthetic and controlled-burn tracks differ in scene construction and in the number of candidate exposures. Fourth, the evaluation considers one frame at a time. A more exposed structure may lie outside the field of view. Important contextual factors, such as wind conditions and suppression activity, may also be only partially visible. These limitations arise from the single-view setting itself and would persist even with a stronger model. Finally, we use one prompt and response schema throughout the experiments. The sensitivity of the results to alternative prompts and response formats has not been evaluated.

These limitations point to several directions for future work. Prior work has shown that fine-tuning VLMs on labeled fire imagery can improve scene understanding [7], while synthetic data can help address limited training data in built environment applications [15]. The synthetic wildfire dataset could therefore be used for model fine-tuning. The dual-track protocol could then evaluate whether these improvements transfer to real wildfire imagery. Future work should also include additional real-burn datasets, evaluation across a broader range of models, and analysis of how wind and weather conditions affect fire-threat assessment.

## VI. Conclusion

This paper contributes a dual-track diagnostic for primary structure threat identification under limited real WUI fire observations. Under zero-shot prompting, Qwen3-VL-8B-Instruct achieved 94.0% accuracy on valid synthetic outputs and 69.2% on valid controlled-burn outputs, compared with a 16.0% random-selection baseline for the latter. Because the tracks differ in scene construction and dataset scale, with 1,824 synthetic images from 491 scenarios but only 45 sampled controlled-burn frames, their results should be interpreted separately.

The synthetic track supports stress testing across varied WUI fire-threat scenarios, and the model usually identified the intended target when fire, fuel, barrier, and structure cues were clear. Controlled-burn errors, however, suggest that the VLM's spatial understanding and threat assessment remain limited in complex or ambiguous real-fire scenes. Accuracy also varied across camera sources that differed in image quality, viewpoint, and scene complexity. Future work could use the synthetic data for VLM post-training and evaluate the adapted models on larger and more diverse real WUI fire datasets.